\documentclass[A4paper, 11 pt, conference]{ieeeconf}  
\IEEEoverridecommandlockouts       
\usepackage{geometry}
\usepackage{graphics} 
\usepackage{epsfig} 
\usepackage{mathptmx} 
\usepackage{times} 
\usepackage{amsmath} 
\usepackage{amssymb}  
\usepackage{subfigure}
\usepackage{stfloats}
\usepackage{algorithm} 
\usepackage{algorithmic} 
\usepackage{multirow} 
\begin{document}

\title{\LARGE \bf
Learning Based Adaptive Force Control of Robotic Manipulation Based on Real-Time Object Stiffness Detection}

\author{Zhaoxing Deng,  Xutian Deng and Miao Li
\thanks{
  Zhaoxing Deng is  with Wuhan Cobot Technology Co., Ltd., Hubei, China, {\tt\small dengzhaoxing@cobotsys.com}. Xutian Deng and Miao Li is with Department of Mechanical Engineering, Wuhan University, Hubei, China.
}}

\maketitle 
\thispagestyle{empty}

\begin{abstract}
Force control is essential for medical robots when touching and contacting the patient’s body. To increase the stability and efficiency in force control, an adaption module could be used to adjust the parameters for different contact situations. We propose an adaptive controller with an Adaption Module which can produce control parameters based on force feedback and real-time stiffness detection. We develop methods for learning the optimal policies by value iteration and using the data generated from those policies to train the Adaptive Module. We test this controller on different zones of a person’s arm. All the parameters used in practice are learned from data. The experiments show that the proposed adaptive controller can exert various target forces on different zones of the arm with fast convergence and good stability.

\emph{Index Terms}—Force Control, Adaptive Control, Stiffness Detection, Value iteration.
\end{abstract}

\section{INTRODUCTION}
\begin{figure}[t]
	\centerline{\includegraphics[width=1\columnwidth]{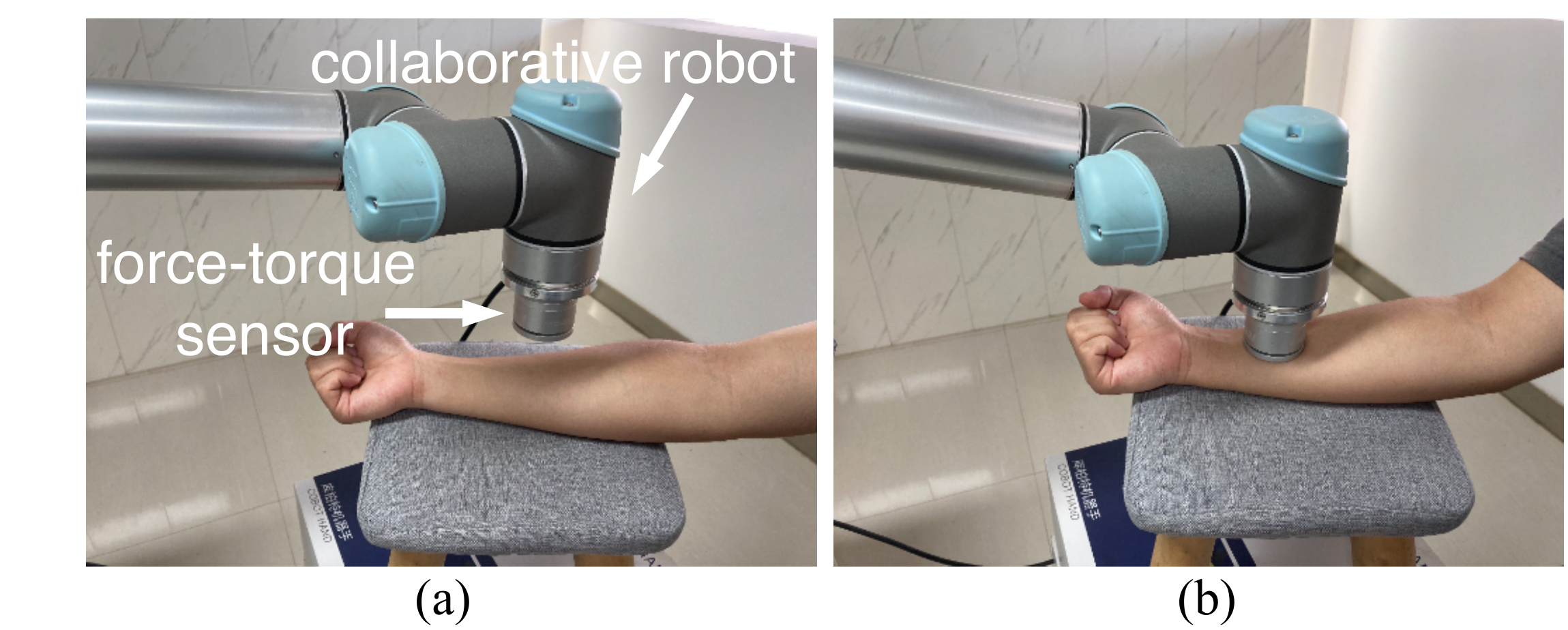}}
	\caption{A typical scenario for contact tasks. A robot is equipped with a force-torque sensor at the end-effector. (a) The robot end-effector is moving downwards. (b) The robot end-effector contacts a person's arm with certain force.}
	\label{Experiment-Preface}
\end{figure}

Contact is everywhere. To accurately control the exerted force when contacting with objects is crucial for medical robots and human-robot interaction. In a typical scenario of surgical applications, a robot can carry a medical device touching and scanning along a patient’s skin (Fig. \ref{Experiment-Preface}), moving to the inspection position under the guidance of doctors, and incising tissues in the body. Force control plays a key role in those tasks. The robot should contact gently with different areas of the body and apply reference forces precisely. This requires the implemented controller to adjust to various contact environments and converge smoothly to the reference value.

Traditional force control methods include PI force control and impedance control. PI force control directly achieves target force through force feedback and PI control law \cite{b1}. It is simple and efficient, but it needs tuning the fixed parameters for each scenario. It also neglects the interaction between robots and environment; thus, it lacks robustness and may cause instabilities and oscillations at contact. Impedance control indirectly control the interaction force between robots and environment through impedance control law \cite{b2}. It simulates interaction dynamics as a mass-spring-damper system and adjusts interaction forces and motions. It is suitable for applying a compliant behavior through contact, so it's widely used in polishing\cite{a1}, grasping\cite{a2} and coating tasks\cite{a3}. But lacks the ability to precisely track forces due to partial knowledge of the environment.  

There are many adaptations of those classical methods. An adaptive controller based on stiffness detection and force feedback for grasping object task is proposed in \cite{b3}. It calculates the stiffness of the object and adjusts the force feedback gain with piecewise linear functions. However, it only focusses on grasping mental objects which have very small deformations. An impedance controller for force-tracking by adapting the control gains from estimated environment stiffness is described in \cite{b4}. The controller uses analytical solutions from LQR to calculate the control gains and EKF to estimate the environment stiffness. That method needs an accurate modelling of the robot dynamics as well as predefined parameters, thus lacks robustness for various types of environments.

One method to learn from the environment and improve the force adaptation is by using Dynamical Systems (DS) framework. In \cite{b5}, an adaptive motion planning approach is used on impedance-controlled robots for polishing and pick-and-place tasks. Two types of parameterized time-independent dynamical systems (DS) are used for motion generation in different tasks. The parameters of the DS will be updated to adapt the robot’s motion to human interactions.  In \cite{b6}, the time-independent DS-based control framework is extended to adjust both the trajectory and the force when preforming contact tasks. The robot’s nominal DS is used to generate the desired motion and contact force. In \cite{b7}, a state-dependent force correction model is proposed to improve force tracking accuracy. Those methods needs to learn the DS profile of the nonlinear surface. 

Another method to enable robust physical iteraction with the environment is to learn the contraints on the manipulated objects. In \cite{a4}, a manifold learning approach is used to encode the Constrained Object Manipulations (COM) tasks. This achieves task-consistent adaptation based on an object-level impedance controller.
\begin{figure}[b]
	\centerline{\includegraphics[width=1\columnwidth]{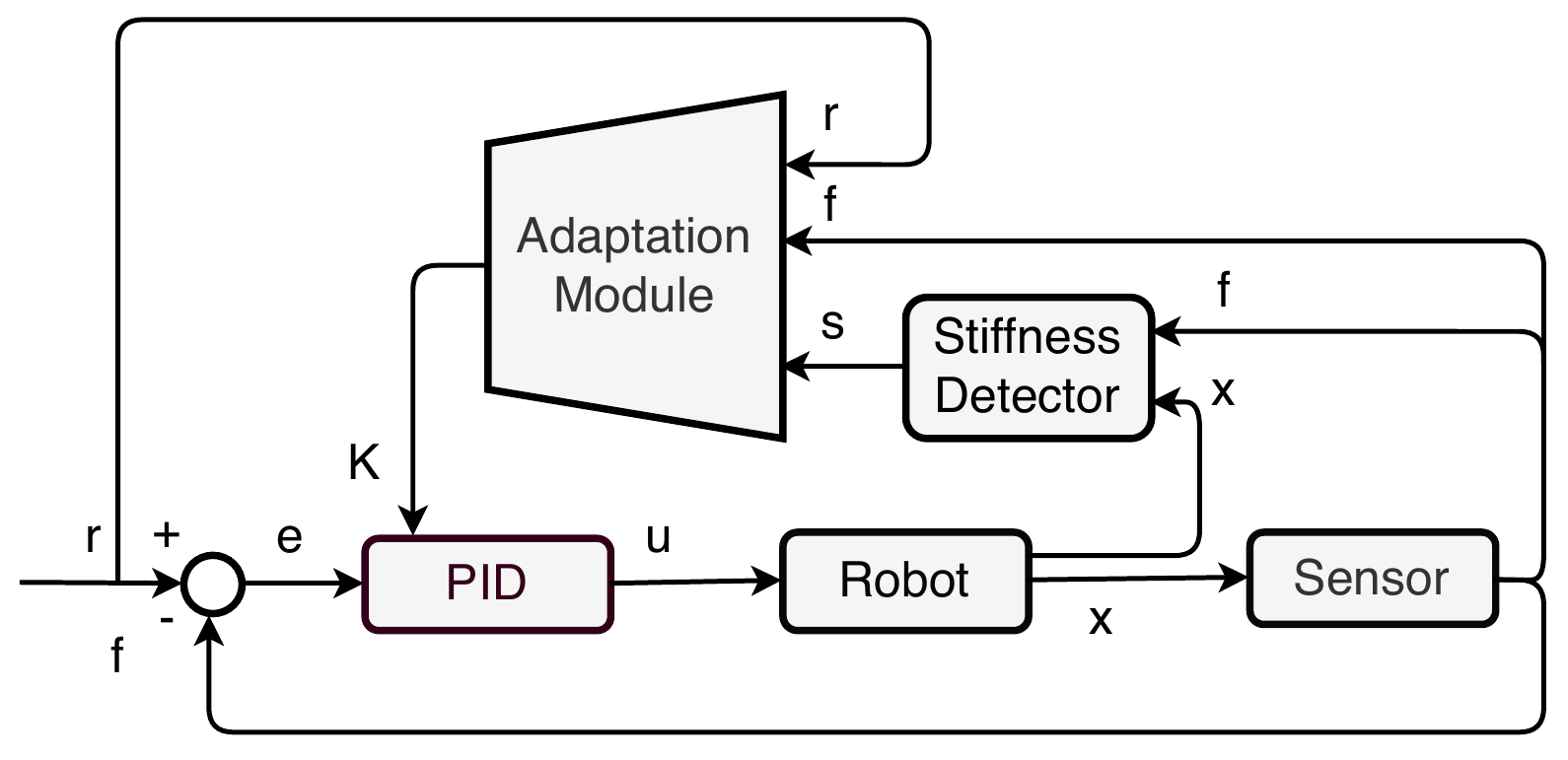}}
	\caption{The block diagram of the adaptive controller}
	\label{Adaptive-Controller}
\end{figure}
The other method for motion and force adaptation is by using Reinforement-Learning based controller. In \cite{b8} the Rapid Motor Adaptation algorithm (RMA) is proposed to allow legged robots move stably in difficult and changing terrains with changing payloads by real-time online adaptation. RMA has two subsystems: a base policy and an adaptation module. Those components are trained completely in simulation. The adaptation module is the key component of RMA which predicts the extrinsic vector using recent history of robot’s states and actions. The trained controller can be deployed directly on the A1 robot without any fine-tuning. 

DQN \cite{b9} and SAC \cite{b10} are two popular model-free learning methods in reinforcement learning. However, if we have a good model of the dynamics, we can adopt model-based learning such as value iteration \cite{b11} to effectively derive the optimal policy for discrete-time non-linear dynamic system.

\begin{figure}[t]
	\centerline{\includegraphics[width=1\columnwidth]{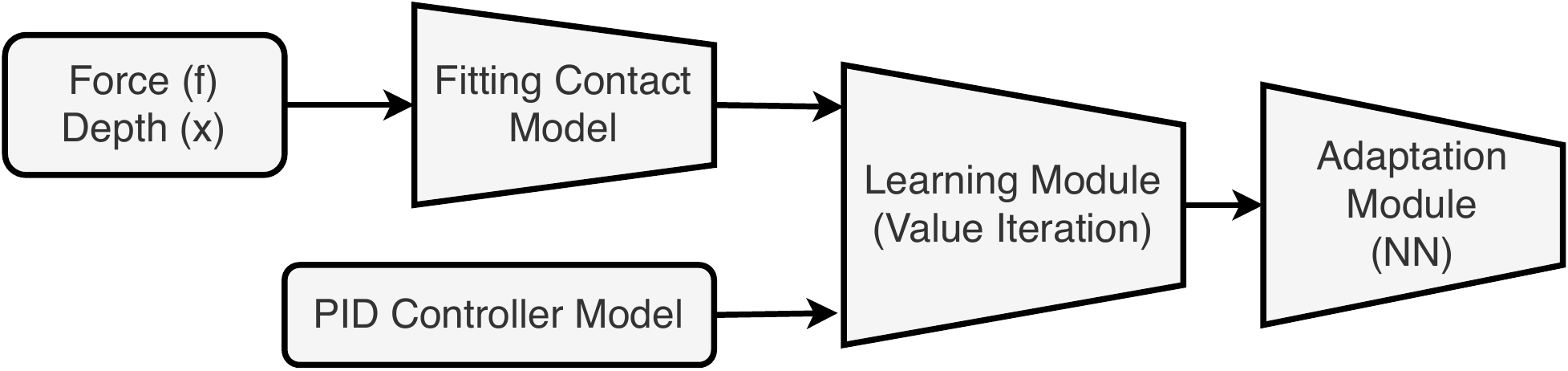}}
	\caption{The process for training the Adaptive Module}
	\label{Learning-Model}
\end{figure}

To exploit the idea of learning from environment and training an adaptation module to improve versatility and stability, we propose an adaptive controller with an Adaptation Module which can adjust the parameters based on force feedback and real-time stiffness detection, as shown in Fig. \ref{Adaptive-Controller}. The Stiffness Detector computes the real-time sitffness of the object. The process for training the Adaptation Module is shown in Fig. \ref{Learning-Model}. We first sample data from different zones of the environment and fit those data to model  local dynamics of those zones. Then we build corresponding dynamic systems and use value iteration to learn the optimal policies for those dynamic models. Finally, we generalize the results to all parts of the environment by training a neural network to learn the relation between environment feature variables and controller parameters. 

We present the design of the controller and training methods in section II. We evaluate the results of  training processes and experiments on different zones of a person’s arm in section III. We conclude with a discussion about the methods and future work in section IV. 

\section{METHODS}

\subsection{Designing the Adaptive Controller}
The adaptive controller is based on a PID controller and an adaptive module which tunes the parameters for the PID controller. As seen from Fig. \ref{Adaptive-Controller}, the reference command $r$ is fed into the system. The error $e = r-f$  is used as the input to the standard PID controller. The controller output $u$ is then used to drive the robot towards the reference force 
\begin{equation}
u = K_p   e + K_i  \int e \ dt + K_d  \frac{de}{dt} \label{pid}
\end{equation}
where $K_p$, $ K_i $, $K_d$ are parameters for the controller.

In a position-controlled robot, $u$ can be the displacement within a control period. $x$ is the position of the robot end-effector in Cartesian space. The force-torque sensor at the end-effector reads the current force value $f$ and feeds that value to the Stiffness Detector as well as the Adaptation Module. The Stiffness Detector uses the current force value and position to calculate the stiffness of the object. Notice stiffness here is defined as the slope of force-deformation curve and thus represents the local property of the material. The equation to calculate the real-time stiffness is 
\begin{equation}
s = \frac{f_c - f_l}{dx_l} \label{stiffness}
\end{equation}
where $f_c$ is the current force value, $f_l$ is the force value in last control period and $dx_l$ is the displacement in last control period. 

The output $s$ is then passed to the Adaptation Module. The Adaptation Module uses the reference value $r$, the current force value$ f $ and current stiffness $ s $ to compute the adjustable parameters $K$ for the PID controller. 

\subsection{Modeling Contact Dynamics from Data}
We model the dynamic dynamics for using dynamical systems in model-based learning. We focus on the relationship between contact force and pressing depth (deformation) of a person’s arm. We gather the data from some specific zones of the arm and generate approximation of dynamic model for other  zones. From observation, when the pressing depth increases, the contact force as well as stiffness is also increasing smoothly. Thus, we can choose general exponential functions to fit the data
\begin{equation}
y = f(x) = a \exp (-b x) + c \label{model}
\end{equation}
where $a$, $b$, $c$ are parameters in the exponential function.

We can use non-linear least squares to fit exponential functions to those curves.

\subsection{Learning Optimal Policies using Value Iteration}
Now that we obtain environment dynamic models and we know the output equation of the PID controller, we can design the dynamic systems as
\begin{equation}
\frac{dx}{dt} = K_p (f_r - f(x)) \label{dynamics}
\end{equation}
where $x$ is the state variable representing the depth of the deformation, $K_p$ is the parameter in the PID controller, $f_r$ is the refence value and $f$ is the function of contact dynamics. Notice that for efficiency, we just include $K_p$ as the parameter without $ K_i$ or $K_d$ terms.

To learn the optimal policies given the dynamical systems and considering the discrete nature of real implementation of control program, we use value iteration algorithm. It’s important that we choose $K_p$ as the input to the system. The cost function is chosen to be in quadratic form 
\begin{equation}
\min_{\pi(x)}  \ \int_{t_0}^{\infty}  [a (f_r - f(x))^2 + b  (K_p)^2 ] \,dt\   \label{cost}
\end{equation}
which pushes the system towards the equilibrium with a balance between speed and input magnitude. 
\subsection{Training the Adaptive Module from data}
We learn the $ K_p-x $ policies as the results from value iteration. Since each $x$ relates to force $f$ and its derivative $\frac{df}{dx}$ from the dynamic models, we can gather those discrete $K_p-x$ pairs and rewrite them to the corresponding $[K_p, f, \frac{df}{dx}] $ pairs. Here we assume that every area of the arm has similar dynamic property and $[f, \frac{df}{dx}] $ pair can fully capture that property at any pressing depth. Although we just sample data from some specific zones of the arm, we can generalize the learned policies to other areas of the arm given the feature $ [f, \frac{df}{fx}]$. Since we treat reference value $r$ as an independent environment variable and we learn the policies for various $r$, we also consider $r$ as one input feature. Thus, our goal is to learn the function from input feature vector $[r, f, \frac{df}{dx}]$ to $K_p$.

Learnning the approximate function from input features to output is a typical supervised learning problem. Consider the dimension of inputs and output as well as the amount of data, we choose a 3-layer fully connected neural network as the Adaptation Module architecture and train this neural network on input feature vectors and $K_p$.

\subsection{Building a Hybrid Dynamical System}
\begin{figure}[t]
	\centerline{\includegraphics[width=1\columnwidth]{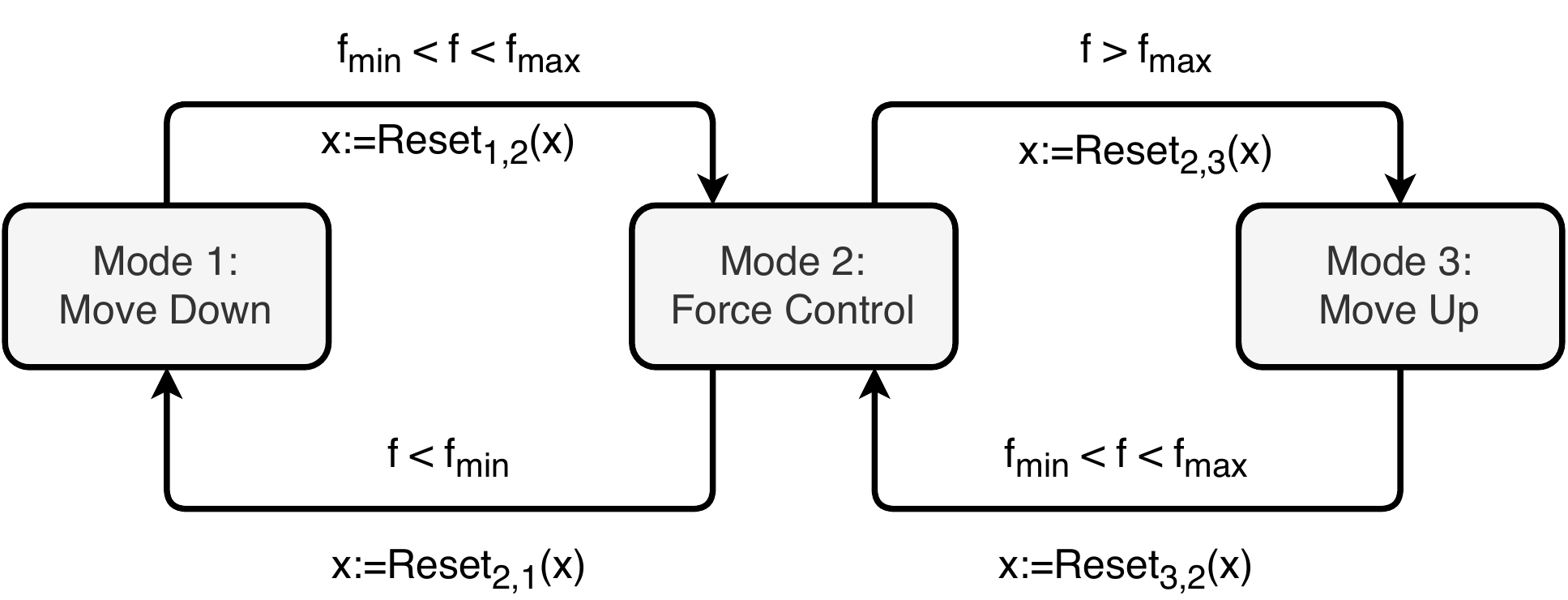}}
	\caption{The hybrid dynamical systesm for the control task}
	\label{Hybrid-Dynamics}
\end{figure}
In practice, we build a hybrid dynamical system for the whole control task, as shown in Fig. \ref{Hybrid-Dynamics}. The control system consists of three modes:
\begin{itemize} 
 \item Mode 1:  Moving downwards when contact force is smaller than the minimum gate $f_{min}$; 
 \item Mode 2:  Using adaptive controller to exert the reference value; 
 \item Mode 3:  Moving upwards when contact force is larger than the maximum gate $f_{max}$. 
\end{itemize}
  The outputs will be reset when transiting from one mode to anther mode.
  
\section{EXPERIMENTAL EVALUATIONS}
\begin{figure}[h]
	\centerline{\includegraphics[width=1\columnwidth]{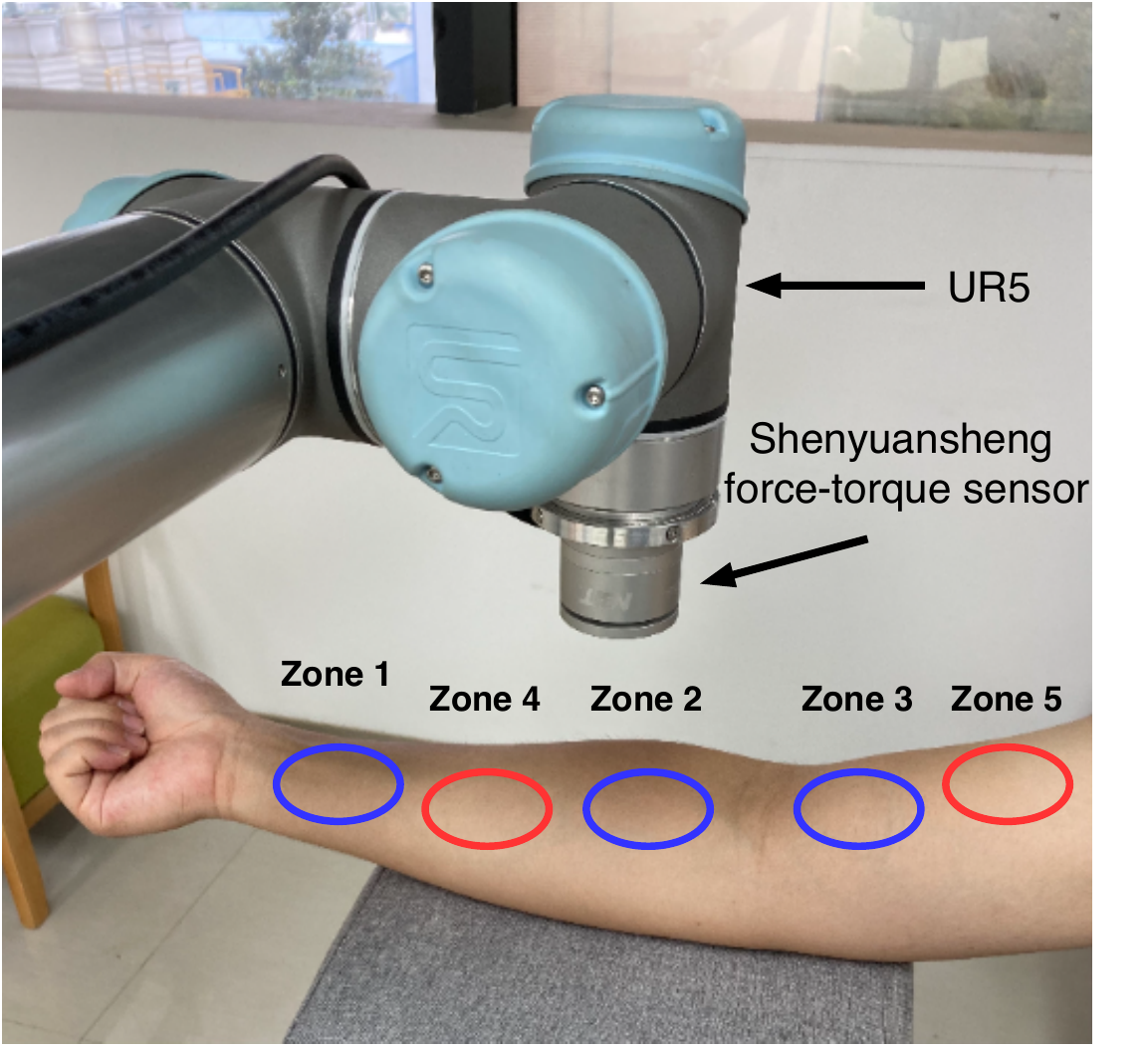}}
	\caption{Blue Zones are the areas to collect force depth data. Red Zones are two other areas on the arm..}
	\label{Experiment-Platform}
\end{figure}
\begin{figure}[t]
	\centerline{\includegraphics[width=1\columnwidth]{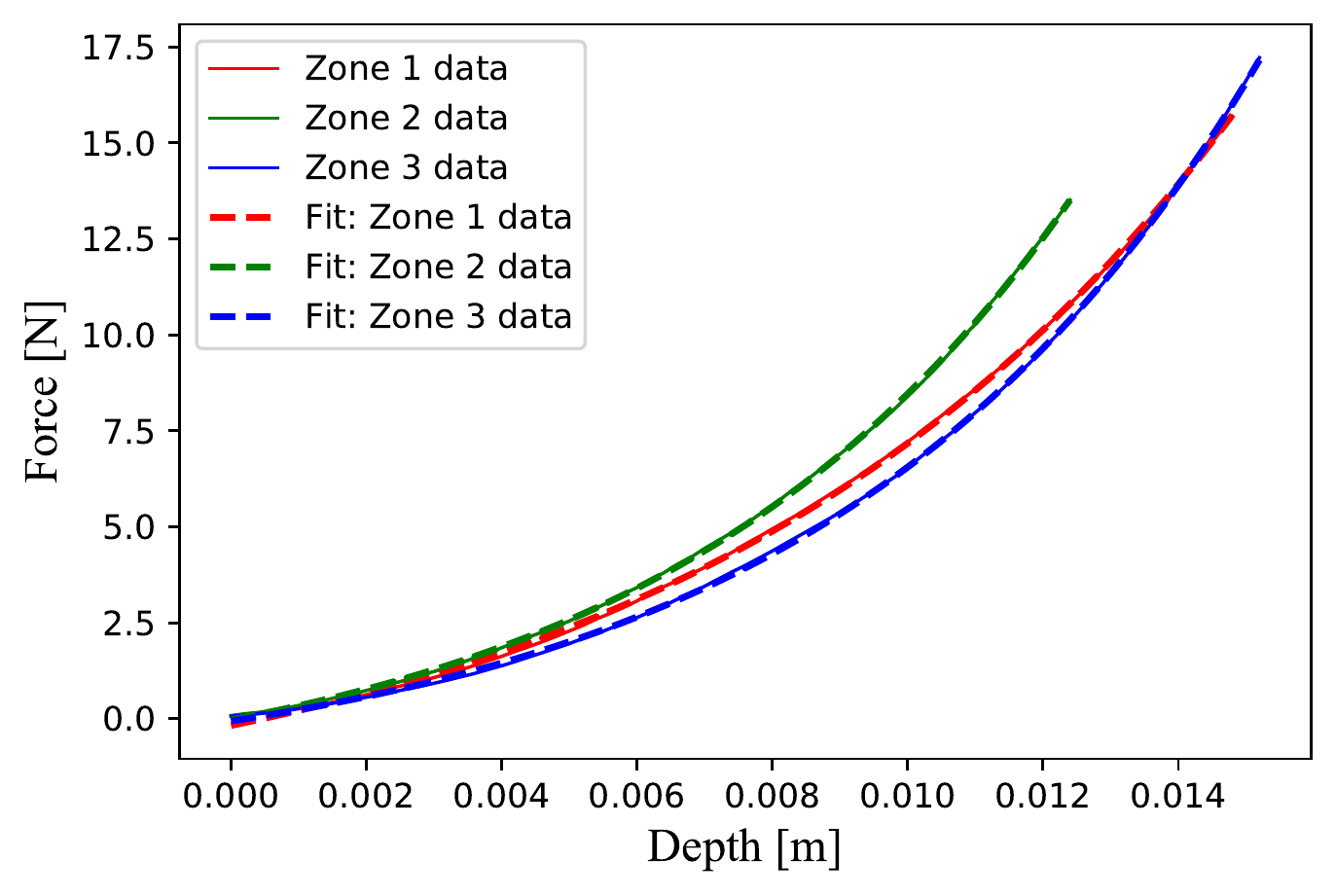}}
	\caption{The solid lines represent original data. The dashed lines represent the fiited functions.}
	\label{Force-Depth}
\end{figure}
\begin{figure*}[h]
	\includegraphics[width=1\textwidth]{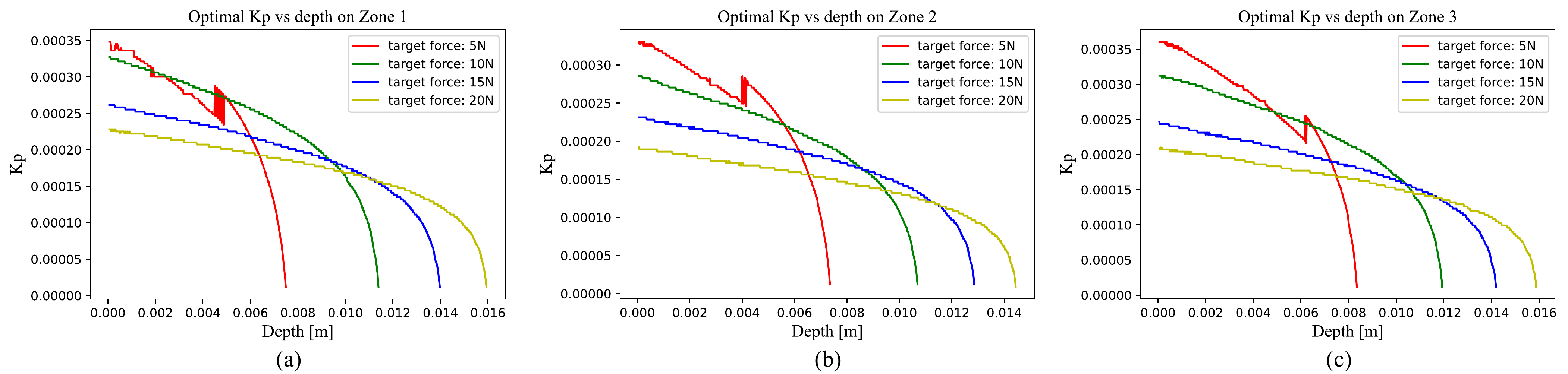}
	\caption{The $K_p - x$ policies for each Zone at various reference values. }
	\label{kp-graphs}
\end{figure*}
In this section, we discuss the processes of training and experiment results. We evaluate our adaptive controller on a UR5 robot arm, as shown in Fig. \ref{Experiment-Preface}. A Shenyuansheng force-torque sensor is attached to its end-effector. The control program is run on CobotSys 3.0 platform, which is an industrial robot operating system developed by Wuhan Cobot Technology Co., Ltd. The control loop rate for this program is $100 Hz$.

\subsection{Fitting the Data for Dynamic Models }
We collect data from three different Zones of a person’s arm shown as the blue circles in Fig. \ref{Experiment-Platform}. We programmed the robot end-effector to move straightly downwards at steady step size until the force being sensed reaches a certain amount. To compensate for the uncertainties in each experiment, we record 10 sets of data on each Zone and take their mean value. The relationships of force and depth at those three Zones are plotted as solid lines in Fig. \ref{Force-Depth}.

Different Zones have different force-depth curves, but they all have similar shapes. Each curve is ascending smoothly with increasing slope. We fit those curves by exponential functions, which perfectly match the data (dashed lines in Fig.\ref{Force-Depth}).

\subsection{Training the Adaptation Module}
To build the dynamic systems and implement value iteration algorithm, we use Drake toolbox \cite{b12}. The state variable $x$ is discretized to be between $[0, 0.02]$ with 1001 steps. The input variable $K_p$ is discretized to be between $ [0, 1] $  with $1000$ steps. The parameters in cost function \eqref{cost} are set to $a = 1$, $ b = 40$. The time step is set to $0.03s$. We choose reference value $r$ in the set $ [4, 24]$ with step size 0.5. The optimal policies $K_p-x$ are computed for various values $ r $ from the three dynamic systems. 
Fig.\ref{kp-graphs}  shows $K_p-x$ policies for three Zones with $r$ ranges from $5 N$ to $20 N$. It’s easy to see that all $K_p-x$ functions have similar shapes. When $ r $ increases, the curves will be flatter, meaning $K_p$ will starts smaller and moves slower along the $x$. Notice that when the reference value $r$ is very small, there will be a little twist on the graph. That is the result of balancing between convergence rate and input magnitude and can be neglected.  The results from value iteration are compatible with our intuition and suggest that the similar policy could be generalized to every other  zone of the arm.

We build a 3-layer neural network for the Adaptation Module with layer sizes of $[6, 3, 1]$. The activation functions of all the nodes are ReLu functions. We train the Adaptation Module using surpervised learning with input feature $[r, f, \frac{df}{dx}]$ and output label $K_p$. We use Adam optimizer  to minimize MSE loss. We run the optimization process for 200 epoches with a learning rate of 1e-4 and use a batch size of  64 split up into 4 mini-batches.
\subsection{Testing on the arm}
\begin{figure}[h]
	\centerline{\includegraphics[width=1\columnwidth]{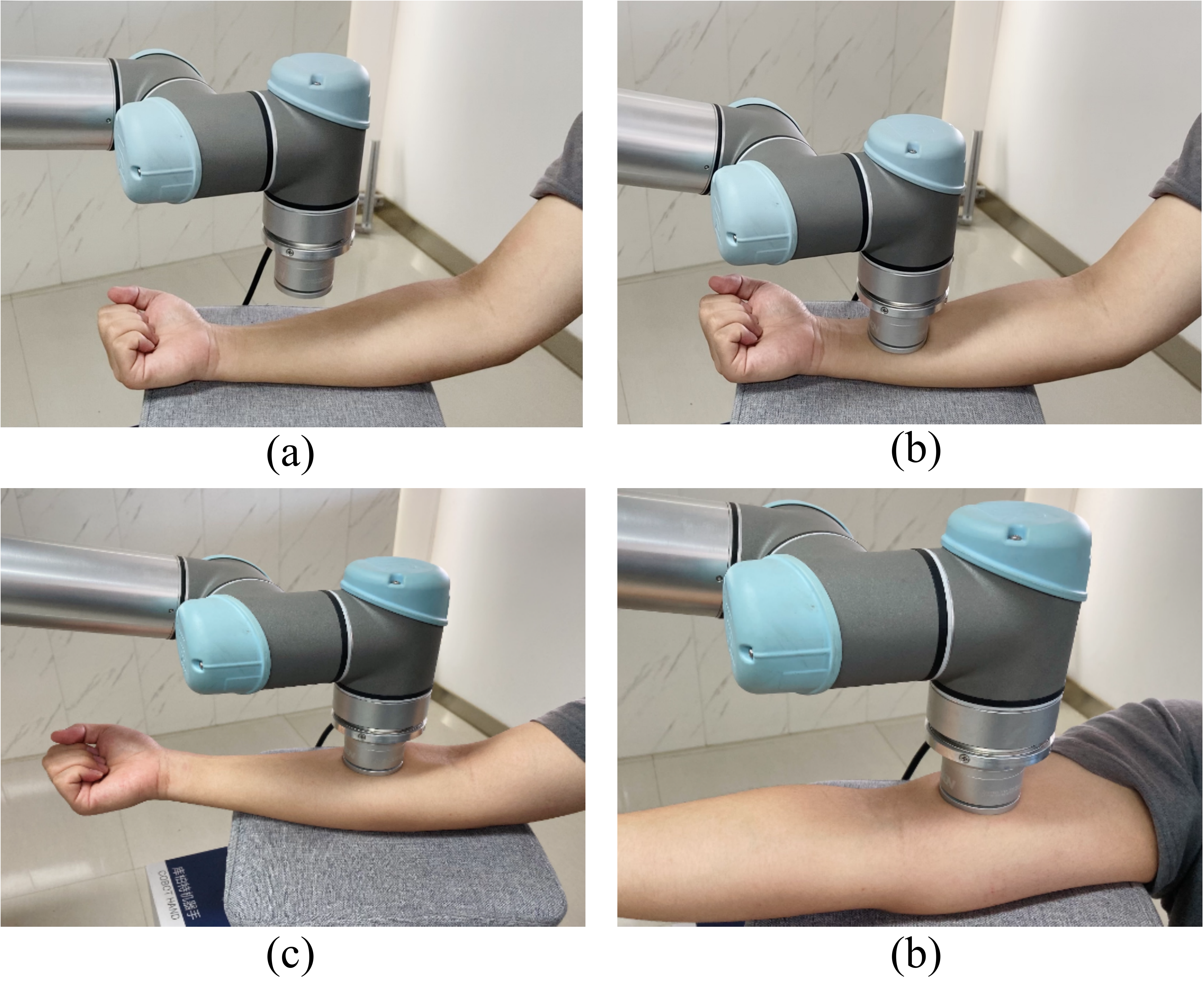}}
	\caption{Test the controller on different Zones of the arm. (a) The robot end-effector is moving straghtly downwards. (b) (c) (d) show the end-effector contacts on different Zones.}
	\label{Experiment-Snapshot}
\end{figure}

We test our adaptive controller on 5 Zones of the arm, as shown in Fig.\ref{Experiment-Platform}. The blue Zones represent the places where we previously get the data for the dynamic models. The red Zones are other two areas on the arm. We range reference values from $5N$ to $20N$. At the beginning of the experiments, the robot's  end-effector will move straghtly downwards, since it's in Mode 1 (Fig.\ref{Experiment-Snapshot}(a)) . After it contacts the arm, it uses adaptive controller to exert reference force value on the arm.

\begin{figure*}[t]
	\centerline{\includegraphics[width=\textwidth]{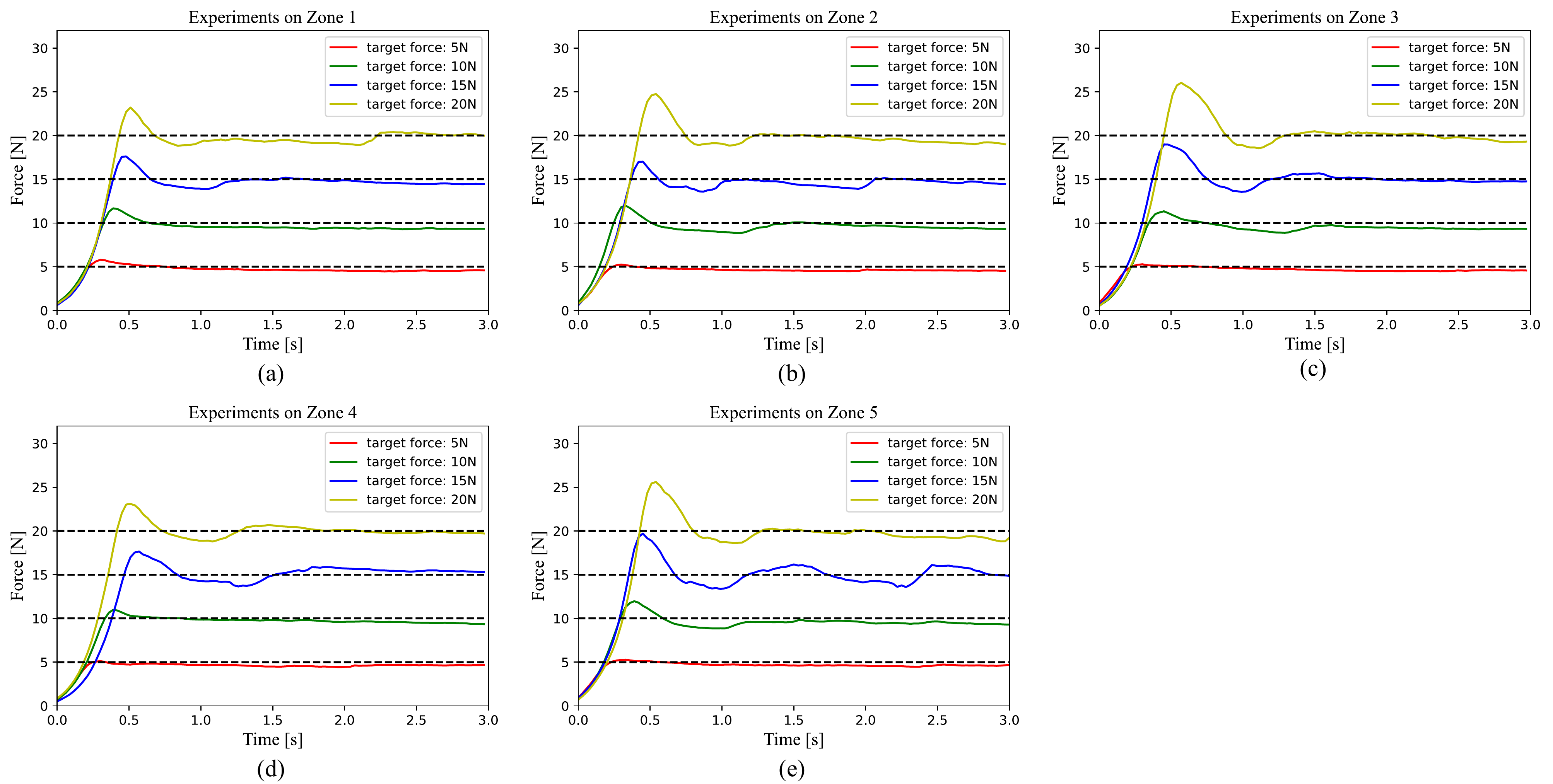}}
	\caption{The results of experiments for each Zone with various reference forces. (a) (b) (c) show the results on previously modeled areas. (c) (d) show results on other areas of the arm.}
	\label{Expriment-Results}
\end{figure*}

Fig. \ref{Expriment-Results} shows the experimental results for each Zone. The convergence time and overshoot will increase as the reference value increases. Typically the convergen time is within $1s$. The overshoot is about $5N$. Due to measurement noises from the force-torque sensor and disterbances of the arm (slight movement of the muscle), there will be some small variations after the curve converges. After all, the controller is very stable. The results show that our controller can converge to the reference value very soon. It can adapt to different contact zones of the arm with good stability and accuracy.
\section{CONCLUSION}
In this paper, we propose an adaptive controller which has an Adaptation Module to adjust parameters for the underlying force controller. The Adaptive Module will take in the reference, current force value and real-time stiffness to output appropriate parameters. The real-time stiffness of the object is evaluated by the Stiffness Detector. We sample data from different zones of the environment and model the local dynamics for each zone. The optimal policies are learned by value iteration from dynamical systems. We then train the Adaptaion Module as a neural network to learn the relations between environment feature vectors and controller parameters. In this way we generalize the results to all zones of the environment. Finally, we build a hybrid dynamical system which combines the adaptive controller and other motion modes for the whole contact task. The experimental results show that our controller can adjust to various environments with good convergence rate, stability, and accuracy. The idea of Adaptation Module and learnining from environment variables is a broad concept which can be extended to a general framework for adaptive controller design. In the future, we will replace the underlying force controller to admittance controller for better stability and consider encoding more environmental variables and constraints in the model to have better performance. We will focus on designing an adaptive controller for contact task on all parts the person’s body.

\section*{ACKNOWLEDGMENT}
The authors would like to acknowledge significant support from Wuhan Cobot Technology Co., Ltd and Wuhan University. The authors especially thank Professor Fei Chen for proofreading and suggestions. The authors also thank Dr. Dong Han and Chu Zhang for discussions about the main idea. In particular, the authors thank Yunchu Zhang for discussions on reinforcement learning techniques and Xiaofeng Chen for helping with the experiments.


\begin{thebibliography}{00}
\bibitem{b1} Kevin M. Lynch and Frank C. Park, Modern Robotics: Mechanics, Planning, and Control, 1st ed, Cambridge University Press, 2017, pp.436-439.
\bibitem{b2} Neville Hogan, "Impedance control: An approach to manipulation,"  in American Control Conference, 1984, pages 304–313. IEEE, 1984.
\bibitem{a1} A. M. Kabir, K. N. Kaipa, J. Marvel and S. K. Gupta, "Automated Planning for Robotic Cleaning Using Multiple Setups and Oscillatory Tool Motions," in IEEE Transactions on Automation Science and Engineering, vol. 14, no. 3, pp. 1364-1377, 2017.
\bibitem{a2} Miao Li, Hang Yin, Kenji Tahara and Aude Billard, "Learning object-level impedance control for robust grasping and dexterous manipulation," 2014 IEEE International Conference on Robotics and Automation (ICRA), pp. 6784-6791, 2014.
\bibitem{a3} Yayun Du, Zhaoxing Deng, et al., "Vision and force based autonomous coating with rollers," 2020 IEEE/RSJ International Conference on Intelligent Robots and Systems (IROS), pp. 9954-9960, 2020.
\bibitem{b3} R. Andrecioli and E. D. Engeberg, "Grasped object stiffness detection for adaptive force control of a prosthetic hand," 2010 3rd IEEE RAS-EMBS International Conference on Biomedical Robotics and Biomechatronics, pp. 197-202, 2010.
\bibitem{b4} Loris Roveda, Niccolo Iannacci, Federico Vicentini, Nicola Pedrocchi, Francesco Braghin, and Lorenzo Moli- nari Tosatti, "Optimal impedance force-tracking control design with impact formulation for interaction tasks," in IEEE Robotics and Automation Letters, 1(1):130–136, 2016.
\bibitem{b5} M. Khoramshahi, A. Laurens, T. Triquet and A. Billard, "From Human Physical Interaction To Online Motion Adaptation Using Parameterized Dynamical Systems," 2018 IEEE/RSJ International Conference on Intelligent Robots and Systems (IROS), pp. 1361-1366, 2018.
\bibitem{b6} Walid Amanhoud, Mahdi Khoramshahi and Aude Billard, “A Dynamical System Approach to Motion and Force Generation in Contact Tasks”, in Robotics: Science and Systems (RSS), 2019.
\bibitem{b7} W. Amanhoud, M. Khoramshahi, M. Bonnesoeur and A. Billard, "Force Adaptation in Contact Tasks with Dynamical Systems," 2020 IEEE International Conference on Robotics and Automation (ICRA),  pp. 6841-6847, 2020.
\bibitem{a4} Miao Li, Kenji Tahara and Aude Billard, "Learning task manifolds for constrained object manipulation", Autonomous Robots 42, 159–174 (2018).
\bibitem{b8} Ashish Kumar, Zipeng Fu, Deepak Pathak and Jitendra Malik, “RMA: Rapid Motor Adaptation for Legged Robots”, in Robotics: Science and Systems (RSS), 2021. 
\bibitem{b9} Volodymyr Mnih, Koray Kavukcuoglu, David Silver, et al., “Human-level control through deep reinforcement learning”, Nature 518, 529–533 (2015).
\bibitem{b10} Tuomas Haarnoja, Aurick Zhou, Pieter Abbeel and Sergey Levine, “Soft Actor-Critic: Off-Policy Maximum Entropy Deep Reinforcement Learning with a Stochastic Actor”, in Proceedings of the 35th International Conference on Machine Learning, PMLR 80:1861-1870, 2018.
\bibitem{b11} Russ Tedrake, Underactuated Robotics: Algorithms for Walking, Running, Swimming, Flying, and Manipulation (Course Notes for MIT 6.832), 2021-6-14.
\bibitem{b12} Russ Tedrake and the Drake Development Team, Drake: Model-based design and verification for robotics, 2019.

\end{thebibliography}
\end{document}